\documentclass[letterpaper, 10 pt, conference]{ieeeconf}  

\IEEEoverridecommandlockouts                              

\usepackage{multirow}
\usepackage{array}
\newcolumntype{C}[1]{>{\centering\let\newline\\\arraybackslash}m{#1}}
\usepackage{graphics}  
\usepackage{epsfig}    
\usepackage{amsmath}   
\usepackage{amssymb}   
\usepackage{epstopdf}
\usepackage[noadjust]{cite} 
\usepackage{float}
\usepackage{rotating}
\usepackage{longtable}
\restylefloat{figure}
\title{\LARGE \bf
Mechanical Design, Modelling and Control of a Novel Aerial Manipulator}

\author{Alexandros Nikou, Georgios C. Gavridis and Kostas J. Kyriakopoulos
\thanks{Alexandros Nikou, Georgios C. Gavridis and Kostas J. Kyriakopoulos are with the Control Systems Lab, Department of Mechanical Engineering, National Technical University of Athens, 9 Heroon Polytechniou Street, Zografou 15780, Greece.
        Email: {\tt\small \{mcp12214,mc08042,kkyria\}@mail.ntua.gr}}%
}

\begin{document}

\maketitle
\thispagestyle{empty}
\pagestyle{empty}

\begin{abstract}
In this paper a novel aerial manipulation system is proposed. The mechanical structure of the system, the number of thrusters and their geometry will be derived from technical optimization problems. The aforementioned problems are defined by taking into consideration the desired actuation forces and torques applied to the end-effector of the system. The framework of the proposed system is designed in a CAD Package in order to evaluate the system parameter values. Following this, the kinematic and dynamic models are developed and an adaptive backstepping controller is designed aiming to control the exact position and orientation of the end-effector in the Cartesian space. Finally, the performance of the system is demonstrated through a simulation study, where a manipulation task scenario is investigated.
\end{abstract}
\section{Introduction}
Aerial manipulation is a new scientific field which has been gaining significant research attention and a wide variety of structures have been proposed in the last years. These manipulation systems possess several features which have lately brought them in the spotlight, with their objective mainly oriented towards performing effectively complex manipulating tasks in unstructured and dynamic environments. Having them include active manipulation as a major functionality, would vastly broaden the applications of these systems, as they move from mere passive observation and sensing to interaction with the environment. Therefore, new scientifically applicable horizons will be introduced related to cooperative manipulation, surveillance, industrial inspections, inspection and maintenance of aerial power lines, assisting people in rescue operations and constructing in inaccessible sites by repairing and assembling. Naturally, both designing and controlling aerial manipulators could be considered as nontrivial engineering challenges.

The first theoretical and experimental results on aerial robots interacting with the environment were developed in \cite{marconi11, marconi22} using a ducted-fan prototype UAV within the framework of AIRobots project. The design of a quadrotor capable of applying force to a wall maintaining flight stability was performed in \cite{sauter}. In \cite{yale1} experimental results with a small helicopter with grasping capabilities were derived, along with the stability proofs while grasping. Several grippers that allow quadrotors to grasp, pick up and transport payloads were introduced in \cite{kumar1}. An implementation of indoor gripping using a low-cost quadrotor has been introduced in \cite{ghadiok}. The authors in \cite{kumar2} addressed the problem of controlling multiple quadrotor robots that cooperatively grasp and transport a payload in three dimensions. Another significant work with cooperative quadrotors throwing and catching a ball with a net was performed in \cite{dandrea}. 

A dexterous holonomic hex-rotor platform equipped with a six DoF end-effector that can resist any applied wrench was proposed in \cite{jiang1}. A system for aerial manipulation, composed of a helicopter platform and a fully actuated seven DoF redundant robotic arm, has been introduced in \cite{kondak}. Another hex-rotor manipulator that consists of three pair of propellers with a two-link manipulator aiming to trajectory tracking control was studied in \cite{kobilarov}. 

More recently, significant experiments using commercial quadrotors equipped with external robotic arms have been conducted in \cite{kim,orsag,cano}.

In this work, a completely novel aerial manipulator is introduced. This could be considered as a small autonomous aerial robot that interacts with the environment via an end-effector by applying desired forces and torques in a 6 DoF task space. The proposed system provides mechanical design flexibility achieved through technical optimization problems. The structural geometric distribution is the outcome of the aforementioned problems with the main goals being oriented towards low body volume, controllability of the system, avoidance of possible aerodynamic interactions and efficiency in performing desired manipulation tasks in dynamic environments. The system is fully integrated as it is not a commercial aerial robot equipped with an external robotic arm, as many of the aerial manipulators mentioned above. The optimal number of thrusters, their positions/orientations and the optimal position of the end-effector on the body structure are defined with respect to the modelling design limitations. Taking all the above into account, the remaining challenge is  
is to actually construct this novel aerial robot.


  

The rest of paper is organized as follows. In Section \ref{sec2} a functional description of the robot and the mechanical design analysis is discussed. A mathematical model that captures the proposed system dynamics and govern the behaviour of the system is derived in Section \ref{model}. Based on this highly nonlinear model, an adaptive backstepping control law is designed in Section \ref{control}. In Section \ref{simul}, simulation results are presented in order to study the performance of the system. Finally, the main conclusions are discussed in Section \ref{conclusions}.
\section{Mechanical Design} \label{sec2}
The overall description of the Aerial Manipulator was based on the idea of designing an aerial robot composed of a set number \textit{n} of similar thrusters and an end-effector, in order to interact with objects in the environment. The exact geometry of the structure will be the result of the analysis of this section. 
\subsection{Principles of the Problem}
\begin{figure}[h]
\centering
\includegraphics[scale=0.64]{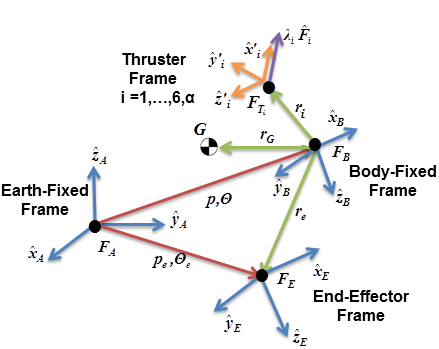}
\caption{Aerial Manipulator Frame Configuration System}
\label{fig:frames}
\end{figure}

Initially, we define the Body-Fixed frame and the End-Effector frame as $F_B= \{\hat{x}_B, \hat{y}_B, \hat{z}_B \}, F_E = \{\hat{x}_E, \hat{y}_E, \hat{z}_E \}$. These frames are attached to the rigid body of the aerial manipulator as in Fig. \ref{fig:frames}. The vectors  $ r_i,r_e \in  \mathbb{R}^3 $ denote the position of each thruster and the position of the end-effector respectively with reference to the Body-Fixed frame. The thruster orientations are given by the unit vectors $\hat{F}_i \in \mathbb{R}^3, \ i = 1,...,n$, the thrust forces are defined as $\lambda_i$ and the propulsion vectors are given by $ \lambda_i \ \hat{F}_i $. At this point, it is assumed that the total system is considered to be a rigid body and, without loss of generality, the End-Effector frame and the Body frame have the same orientation. Thus, the actuation force applied to the end-effector is $F_{act} \big \lvert_B = F_{act} \big \lvert_E  \in \mathbb{R}^3$, where $\big \lvert_B, \big \lvert_E$ denote the expressions to the frames $F_B, F_E$ respectively. The corresponding actuation torque is obtained via the formula $T_{act} \big \lvert_B = T_{act} \big \lvert_E +r_e \times f_e$ where $T_{act} \big \lvert_E = r \times f_e$ is the torque produced by the end-effector. The terms $r,f_e$ denote the displacement vector (length of the lever arm) and the vector force that tends to rotate a gripped object from the end-effector.
\subsection{Forces and Torques}
The forces transmitted essentially through the end-effector are written as
\begin{equation} \label{Fact 1} 
\sum_{i=1}^{n} (\lambda_i \ \hat{F}_i) + W = F_{act} \big \lvert_B
\end{equation}
where $W \in \mathbb{R}^3$ is the vector that corresponds to the total weight of the system. By separating the weight as $w_s = W - n \ w$, where $w \in \mathbb{R}^3$ is the weight of each thruster, \eqref{Fact 1} is modified as 
\begin{equation} \label{Fact 2} 
F \ \lambda +  n \ w + w_s = F_{act} \big \lvert_B
\end{equation}
where $\displaystyle \lambda = [ \lambda_1 \cdots \lambda_n ]^{\tau} \in \mathbb{R}^n$ and $\displaystyle F = [ \hat{F}_1 \cdots \hat{F}_n ]^{\tau} \in \mathbb{R}^{3 \times n}$.

Similarly, the torque from each thruster is $T_i = r_i \times (\lambda_i \hat{F}_i) = \lambda_i S(r_i) \hat{F}_i$. The well-known skew-symmetric matrix $ S(\cdot) \in \mathbb{R}^{3 \times 3} $ is defined as $ a \times b = S(a) \ b $ for the cross-product $ \times $ and any vectors $ a,b \ \in \mathbb{R}^{3} $. The torque due to the weight is calculated as
\begin{equation}
T_W = r_G \times W = \displaystyle \sum_{i=1}^{n} r_i \times w + r_s \times w_s = \displaystyle \sum_{i=1}^{n} S(r_i) w + S(r_s) w_s \nonumber
\end{equation}
where $ r_G $ is the centre of gravity of the system and $r_s$ is the centre of gravity of the system, when omitting the mass of each thruster ($m_{\text{thr}}$). The reaction torque of each thruster is $\tau_i = \mu \ (\lambda \ \hat{F}_i)$ where $\mu $ is a coefficient that represents the relationship between the thrust force and the reaction torque \cite{Padfield}. Therefore, by combining all torques the following equation holds
\begin{multline} \label{Mact 3}
\sum_{i=1}^{n} \left\lbrace  \lambda_i \ S(r_i) \ \hat{F}_i \right\rbrace +\mu \ \sum_{i=1}^{n} \left( \lambda_i \ \hat{F}_i\right) \\ 
+ \sum_{i=1}^{n} S(r_i) \ w + S(r_s) \ w_s = T_{act} \big \lvert_B
\end{multline}
Using the matrices 
\begin{align} \label{eq:simple}
& r = \begin{bmatrix} r_1 & \cdots & r_n \end{bmatrix}^{\tau} \in \mathbb{R}^{3 \times n}, \bar{F} = F_{act} \big \lvert_B-n \ w - w_s  \\
& E(r,F) = \begin{bmatrix} S(r_1) \ \hat{F}_1  & \cdots & S(r_n) \ \hat{F}_n \end{bmatrix} \in \mathbb{R}^{3 \times n} \label{r,F,E}
\end{align}
in \eqref{Fact 2},\eqref{Mact 3} we get
\begin{equation} \label{system}
\begin{cases}
   F \ \lambda = \bar{F} \\
   E \ \lambda = T_{act} \big \lvert_B - \mu \ \bar{F} - \displaystyle \sum_{i=1}^{n} S(r_i) \ w - S(r_s) \ w_s \\
\end{cases}
\end{equation}
By defining the matrix $D(r,F) = \begin{bmatrix} F \\ E(r,F) \\ \end{bmatrix} \in \mathbb{R}^{6 \times n}$, from the system \eqref{system} it is implied that
\begin{equation}\label{Dl=WR}
   D(r,F) \ \lambda = W_R
\end{equation}
where the augmented wrench vector $W_R \in \mathbb{R}^6$ is given by
\begin{equation} \label{eq:WR}
W_R = 
\begin{bmatrix}
   \bar{F} \\
   T_{act} \big \lvert_B -\mu \bar{F} - \displaystyle \sum_{i=1}^{n} S(r_i) w - S(r_s) w_s \\
\end{bmatrix}
\end{equation}
\subsection{Negative Thrust Forces}
It is clear that when solving \eqref{Dl=WR}, the vector that corresponds to the thrust force $\lambda$ can obtain any value in $ \mathbb{R}^6 $. However, the thrusters are optimally designed to produce thrust force towards a specific direction, which we set to correspond to the positive values of $\lambda_i$. In order to alleviate the problem of negative $\lambda_i$, a conservative solution is adopted in this analysis, which is based on the idea of introducing one additional thruster. Thus, \eqref{Dl=WR} is rewritten as
\begin{equation} \label{lt=WR}
\sum_{i=1}^{n} \lambda_i \ t_i = W_R
\end{equation}
where
\begin{equation}
D(r,F) = 
\displaystyle [t_1 \ t_2 \ \cdots \ t_n ] \ \text{and} \ t_i =
\begin{bmatrix}
   \hat{F}_i \\
   S(r_i) \ \hat{F}_i \\
\end{bmatrix} \in \mathbb{R}^6
\end{equation}
for all $i = 1,...,n$. The vector 
\begin{equation} \label{assistive vector}
t_a
=
- \sum\limits_{i=1}^{n} t_i 
=  
\begin{bmatrix}
   -\sum\limits_{i=1}^{n} \hat{F}_i \\[0.8em] 
   -\sum\limits_{i=1}^{n} \left\lbrace  S(r_i) \hat{F}_i \right\rbrace 
\end{bmatrix}
=
\begin{bmatrix}
   \hat{F}_a \\ 
   S(r_a) \hat{F}_a \\
\end{bmatrix}                                                       
\end{equation}
that corresponds to the additional thruster is introduced. Using \eqref{assistive vector}, the position vector $r_a$ and the direction $\hat{F}_a$ of the new thruster should satisfy the equations
$ \hat{F}_a = - \sum\limits_{i=1}^{n} \hat{F}_i \ , \ S(\hat{F}_a) \ r_a = - \sum\limits_{i=1}^{n} \left\lbrace S(\hat{F}_i) \ r_i \right\rbrace$.

\begin{figure}[h]
\centering
\includegraphics[scale=0.66]{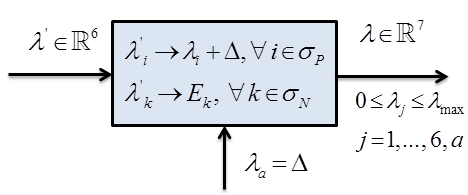}
\caption{Thrust Force Equivalence}
\label{fig:fig_thrust}
\end{figure}

If we assume that \eqref{Dl=WR} results in some negative thrust forces, then the set $ \sigma_{N} = \left\lbrace k: \lambda_k < 0, k = 1,...,6 \right\rbrace  $ denotes the indexes for every negative thrust force and $ \sigma_P = \left\lbrace  1,2,3,4,5,6 \right\rbrace-\sigma_N  $ the corresponding set of positive thrust forces. Observing that $ \lambda_k < 0 \Leftrightarrow (- \lambda_k) > 0, \ \forall \ k \in \sigma_N $, \eqref{lt=WR} can be separated in
\begin{align} \label{eq:proof1}
\sum_{i \in \sigma_P}^{} \lambda_i \ t_i + \sum_{k \in \sigma_N}^{} \lambda_k \ t_k &= W_R \nonumber \\
\Leftrightarrow \sum_{i \in \sigma_P}^{} \lambda_i \ t_i + \sum_{k \in \sigma_N}^{} (-\lambda_k) \ (-t_k) &= W_R
\end{align}
Now, from \eqref{assistive vector} the following can be exported
\begin{equation} \label{eq:proof2}
t_a = - \sum\limits_{i=1}^{n} t_i = -\sum_{i \in \sigma_P}^{} t_i -\sum_{j \in \sigma_N}^{} t_j
\end{equation}
It is obvious that 
\begin{equation} \label{eq:proof3}
-\sum_{j \in \sigma_N}^{} t_j = -t_k-
\sum_{\substack{ j \in \sigma_N \\ j \neq k}} t_j, \ \forall \ k \in \sigma_N
\end{equation}
Combining \eqref{eq:proof2}, \eqref{eq:proof3} we obtain
\begin{equation} \label{eq:proof4}
-t_k = t_a+\sum_{i \in \sigma_P}^{} t_i+\sum_{\substack{ j \in \sigma_N \\ j \neq k}} t_j, \ \forall \ k \in \sigma_N
\end{equation}
By substituting \eqref{eq:proof4} into \eqref{eq:proof1} we have
\begin{align}
\sum_{i \in \sigma_P}^{} \lambda_i \ t_i + \sum_{k \in \sigma_N}^{} (-\lambda_k) \bigg[  t_a+\sum_{i \in \sigma_P}^{} t_i+\sum_{\substack{ j \in \sigma_N \\ j \neq k}} t_j \bigg]  = W_R \nonumber
\end{align}
Defining
\begin{equation}
\Delta = \sum_{k \in \sigma_N}^{} (-\lambda_k) > 0, \ E_k = \sum_{\substack{ j \in \sigma_N \\ j \neq k}} (-\lambda_j) > 0
\end{equation}
and rearranging the terms we result in
\begin{equation} \label{eq:final proof}
\sum_{i \in \sigma_P}^{} \big( \lambda_i + \Delta \big) t_i +
\sum_{k \in \sigma_N}^{} E_k \ t_k+ \Delta \ t_a = W_R
\end{equation}
From \eqref{eq:final proof} the thruster redistribution among all thrusters after adding the new thruster is provided. It has been proven that the issue of negative thrust forces can be alleviated with adding one extra thruster. This equation can be better analysed in Fig. \ref{fig:fig_thrust} in which the thrust redistribution algorithm is depicted. The variables $\lambda', \lambda_i',\lambda_k'$ denote the initial thrust forces and the other variables the thrust force after the redistribution, plus the additional thrust force $ \lambda_a $. Thus, the six thrust forces (not necessary all positive) are equivalent to seven thrust forces, all positive with redistributed thrust forces as in \eqref{eq:final proof}. By using the additional thruster, \eqref{eq:simple}, \eqref{eq:WR} are reformed into
\begin{align}
\bar{F} &= F_{act} \big \lvert_B-(n+1) \ w - w_s \label{WR final} \\
W_R &= 
\begin{bmatrix}
   \bar{F} \\
   T_{act} \big \lvert_B -\mu \ \bar{F} - \displaystyle \sum_{i=1}^{n+1} S(r_i) \ w - S(r_s) \ w_s \\
\end{bmatrix} \label{WR final2}
\end{align}
\subsection{Aerodynamic Interaction}
At this point, the aerodynamic interaction between the operation thrusters is investigated. The aerodynamic effects produced by each thruster, are based on experiments that took place in Control Systems Lab NTUA on a $8\times 4.7SF$ APC propeller accompanied with the Neu Motor NEU 1902/2Y - 2035 motor, which produces at 17550 rpm, a $\lambda_{\text{max}} = 28 N$ thrust force. The surface, corresponding to every thruster, that approximates these effects is described by a third order equation in (SI). By expressing this equation in the thruster frame $F_{T_i} = \{\hat{x}'_i, \hat{y}'_i, \hat{z}'_i\}, i =1,...,6,a$ we get
\begin{align} \label{equations of the aerodynamic effects at 17550}
-0.06 &\leq x'_i \leq 0.91 \\
(y'_i)^2 + (z'_i)^2 &\leq  \left[ -1.1 (x'_i)^3 + 1.56 (x'_i)^2 - 0.3 (x'_i) + 0.11 \right]^2 \nonumber
\end{align}
Hence, the aerodynamic effects of the air flow throughout the rotor are extended from $x=-0.06m$ to $x=0.91m$. The $x$ axis shows the length of the aerodynamic effect of the exit flow. In order to understand this, one should consider the rotor/blade to be positioned at $x=0$. On the other hand, $y$ axis shows the distance of the effect measured from the rotation axis of the blade, where at position $(x=0,y=0.102m)$ (SI) is the blade radius in approximation (because of the existence of an offset). 

An arbitrary point $p = \left[ x \ y \ z \right]^{\tau}$ expressed in $ F_B $ and the corresponding $p'_i =  \left[ x'_i \ y'_i \ z'_i \right]^{\tau}$ expressed in $F_{T_i}$, can be linked together by the equation $p = \{ TR_{F_B}^{F_{T_i}} (r_i,\hat{F}_i) \} \ p'_i$, where $ TR_{F_B}^{F_{T_i}} (r_i,\hat{F}_i) $ is the appropriate homogeneous frame transformation corresponding to the translation and orientation vectors $ (r_i,\hat{F}_i) $. By combining the previous coordinates transformation equation with the constraints \eqref{equations of the aerodynamic effects at 17550}, a set of constraints that can be described in matrix form as $G(r_i,\hat{F}_i,p) \leq 0$, is produced. The distance between two such volumes $i$, $j$ can be defined and evaluated via the optimization problem $(P_1)$ of Table \ref{table:opt_problems}.
\subsection{Design Problem}
\label{Design problem}
Given a particular structure defined by the matrices $(r,F)$, for a set of required actuation forces and torques $(F_{act} \big \lvert_E, T_{act} \big \lvert_E )$ it is necessary to find the associate levels of the thrust forces $ \lambda_i $. Since $ W_R \in \mathbb{R}^6 $, in order for \eqref{Dl=WR} to have a solution for $ \lambda \in \mathbb{R}^n $, the conditions $ \{ \text{rank}(D) =6, n \geq 6 \}$ are required. The rank condition is adequate from a strict mathematical perspective but from a practical point of view, as \eqref{Dl=WR} leads to the thrust forces values  $\lambda \in \mathbb{R}^n$, the sought solutions should not be very sensitive to small deviations. This is partially achieved by using the condition number $ \kappa(D) = \sigma_{\text{max}}(D) / \sigma_{\text{min}}(D) $ where $ \sigma(D) = \sqrt{\text{eig} ( \displaystyle D^{\tau} D)} $ are the singular values of the matrix D, $\text{eig}(\cdot)$ denotes the eigenvalues of a matrix and $\sigma_{\text{max}}(D), \sigma_{\text{min}}(D)$ are the maximum and minimum singular values of the matrix $D$ respectively. Thus, a low condition number $ \kappa(D) \geq 1 $ is required \cite{Lloyd}. Although the condition number is bounded to take feasible values (not equal to zero/infinity) when $\sigma(D) \rightarrow 0 $, the matrix $ D(r,F) $ might be ill-conditioned i.e. $ \text{det}(D(r,F)) \rightarrow 0 $. Thus, $\sigma(D) \geq \epsilon_1 > 0 $. Furthermore, to avoid the fan interaction an other constraint is introduced as $ d_{ij}(r_i,\hat{F}_i,r_j,\hat{F}_j) \geq \epsilon_2 > 0, \ \forall i,j =1,2, \dots,n,\alpha  $. Note that, similarly to $(P_1)$, the position $r_e$ should be introduced to the design problem as the intersection avoidance between a sphere (with radius $R_e$) that encloses the end-effector, and the thrusters. This sphere, when expressed in the End-Effector frame $ F_{E} $, is given by $(x'_e)^2 +(y'_e)^2 + (z'_e)^2 \leq  R_{e}^{2}$. Therefore,  the constraint associated with the end-effector is $ d_{ei}(r_e,r_i) \geq R_e > 0, \ \forall i =1,2, \dots,n,\alpha$. An optimization is also required to minimize the volume of the system, by using the norm $J(r)= \| r \|_{2}$. Taking all the above into consideration, the design problem is essentially recast to the optimization problem $(P_2)$ from Table \ref{table:opt_problems}. The optimization parameters are chosen as $ K = 5, \ \epsilon_1= 10^{-3}, \ \epsilon_2 = 10^{-2} \ m, \ R_e=10^{-2} \ m$.  
\begin{table}[h]
\begin{center}
\begin{tabular}{|C{0.6cm}|m{6.7cm}|}
\hline
{\multirow{3}{*}{ $(P_1)$ }} & $\qquad \qquad \ d_{ij} (r_i,\hat{F}_i,r_j,\hat{F}_j) = \underset{p_i,p_j}{\operatorname{min}} \| p_i - p_j \|$  \\

  & $\qquad \qquad s.t. \ \ \ G(r_i,\hat{F}_i,p_i) \leq 0 $ \\

  & $ \qquad \qquad \qquad \ G(r_i,\hat{F}_i,p_j) \leq 0$ \\        
\hline
{\multirow{7}{*}{ $(P_2)$ }} & $\qquad \qquad \qquad \qquad \underset{r,r_e,\hat{F}}{\operatorname{min}} \ J(r)$  \\
    & $\qquad \qquad s.t. \ \ \ \sigma(D) \geq \epsilon_1$ \\
    & $\qquad \qquad \qquad \ \ d_{ij} \geq \epsilon_2, \ \forall \ i,j =1,2, \dots,n,\alpha $ \\
    & $\qquad \qquad \qquad \ \ d_{ei} \geq R_e, \ \forall \ i=1,2, \dots,n,\alpha $ \\
    & $\qquad \qquad \qquad \ \ \displaystyle \hat{F}_a = - \sum\limits_{i=1}^{n} \hat{F}_i $ \\
    & $\qquad \qquad \quad \quad \ \ \displaystyle S(\hat{F}_a) \ r_a = - \sum\limits_{i=1}^{n} S(\hat{F}_i) \ r_i $ \\
    & $\qquad \qquad \qquad \ \ 1 \leq \kappa(D) \leq K $ \\                
\hline
\end{tabular}
\end{center}
\caption{Optimization Problems}
\label{table:opt_problems}
\end{table} 
\subsection{Solving the Optimization Problem}
It should be noted that when solving the optimization problem $(P_2)$, each time the inner problem $(P_1)$ should be solved. There are 45 decision variables of the optimization problem, which correspond to the seven position vectors ($r_i$) of the thrusters, the position vector ($r_e$) of the end-effector and the direction vectors ($\hat{F}_i$) of the seven thrusters. This issue, entails the necessity of solving 28 optimization problems for each evaluation attempt of the outer problem $(P_2)$.

The inner problem, that refers to the avoidance of the fan interaction, is smooth but in terms of the outer problem $(P_2)$ is nonsmooth and nonlinear. The objective function and the constraints of the problem $(P1)$ are continuous and this problem, according to the inputs, has one and only one global minimum. Using the appropriate rotation and transformation matrices, the $(P1)$ was solved by the active-set strategy \cite{activeset1},\cite{activeset2}. On the other hand, the design problem $(P_2)$ has nonsmooth, discontinuous and nonlinear inequality constraints, but smooth objective function. Consequently, a non-gradient-based methodology that searches disjoint feasible regions, is utilized. For the pre-search of the design space, a Latin Hypercube (LHS) \cite{LHS} was chosen, in order to ensure that the points are distributed throughout the search space. The Latin Hypercube sampling is known to provide better coverage than the simple random sampling \cite{LHcomp}. Following this, a Generalized Pattern Search (GPS) direct search algorithm \cite{GPS1},\cite{GPS2} was used. 

\begin{figure}[h]
\centering
\includegraphics[width=92mm,height=49mm]{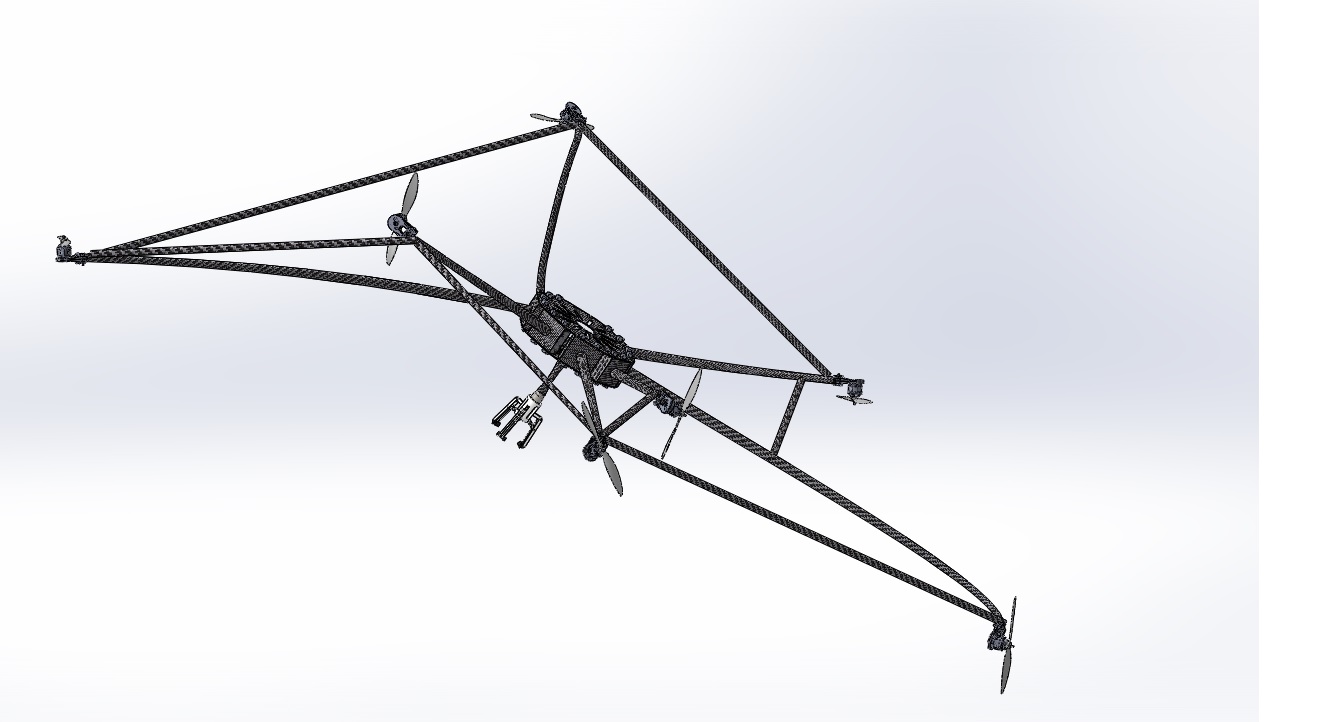}
\caption{Aerial Manipulator 3D Caption of the Framework}
\label{fig:Solidworks}
\end{figure}

The thrust force ($ \lambda $) and the momentum (Q) can be calculated from \cite{Martinez},\cite{Prouty} as
\begin{equation}\label{Q,T}
\begin{cases}
Q =  \pi \rho \ C_Q \ R^5 \ \Omega^2 \\
\lambda = \pi \rho \ C_{\lambda} \ R^4 \ \Omega^2 \\
\end{cases} \Leftrightarrow \ \ Q = \frac{C_Q}{C_{\lambda}} \ R \ \lambda
\end{equation}
where the term $\frac{C_Q}{C_{\lambda}} \ R $ corresponds to the coefficient $ \mu $, $ R $ is the radius of the rotor and $ \rho, \Omega $ denote the air density and the rotational speed of the rotor respectively. Applying a combination of the Blade Element Theory \cite{Prouty} and the Momentum Theory \cite{Padfield}, using the modified versions proposed in \cite{Martinez} and invoking the experimental results extracted by our lab on the APC propeller, it was calculated that $C_{\lambda}=0.008, C_Q=0.0095, \mu = 0.1473, R = 0.124m$. By solving the optimization problems, with the results depicted in Table \ref{table_parameters}, the matrix $D(r,F)$ is full rank and using \eqref{Dl=WR}, the thrust forces can be calculated as $\lambda = D^{-1} \ W_R$. All the constraints were satisfied and a low volume body structure with condition number $ \kappa(D) = 3.36 $ resulted. The wrench vector $W_R$ can be determined by substituting the desired actuation forces/torques ($F_{act} \big \lvert_E$, $T_{act} \big \lvert_E$) in \eqref{WR final2}. The maximum thrust force and torque which can be applied from the system are $\lambda_{\text{max}}= 28 N$ and $ 3 Nm $ respectively. The values of the components, proposed for the Aerial Manipulator, are the following: the motor and the propeller $(0.12 kg)$, the frame $(0.66 kg)$, the battery $(0.25 kg)$ and the electronic components $(0.15 kg)$. The total mass of the proposed system is $m = 1.90 kg$. Ultimately, the production of a carefully studied framework (Fig. \ref{fig:Solidworks}) was achieved by using the 3D CAD Package SolidWorks. Using this Package, the system parameter values of the Table \ref{table_parameters} have been evaluated.
\begin{table}[h]
\begin{center}
\begin{tabular}{|C{0.6cm}||C{1.8cm}||C{3.44cm}||C{0.56cm}|}
\hline
Param.  & Description & Value & Units \\
\hline \hline
$ m $  & Total Mass & 1.90 & $ kg $  \\
\hline
$ m_{thr} $ & Thruster Mass & 0.12 & $ kg $ \\
\hline
$ I_G $ & Moment of Inertia Tensor & $ \left[ \begin{smallmatrix} 0.3488 & 0.0683 & -0.0457 \\ 0.0683 & 0.1588 & 0.0144 \\ -0.0457 & 0.0144 & 0.4081 \\ \end{smallmatrix} \right] $ & $ kg \ m^2 $  \\
\hline
$ r_G $ &  Centre of Gravity Position & $ [ 0.0737 \ 0.0083 \ -0.0781]^{\tau} $ & $ m $ \\
\hline
$ r_e $ &  End-Effector Position & $ [ -0.23 \ 0.015 \ 0.23]^{\tau} $ & $ m $ \\
\hline
$ r_s $ & Centre of Grav. from \eqref{Mact 3} & $ [ 0.1267 \ -0.0052 \ -0.1900]^{\tau} $ & $ m $ \\
\hline
$ J(r) $ & Total Structure Volume & 1.80018 & $ m^3 $ \\   
\hline
$ g $  & Gravitational Acceleration & 9.81 & $ m/s^2 $  \\
\hline 
{\multirow{7}{*}{ $ r_i $ }} & {\multirow{3}{*}{ }} & $ r_1 = [ 0.43 \ -0.15 \ -0.44]^{\tau} $ & {\multirow{7}{*}{ $ m $ }} \\

 &  & $ r_2 = [ 0.08 \ -0.22 \ -0.14]^{\tau} $ &  \\

 &  & $ r_3 = [ 0.1 \ -0.9 -0.2]^{\tau} $ &  \\

 & Thruster Positions & $ r_4 = [ -0.34 \ 0.25 \ 0.006]^{\tau} $ &  \\

 &  & $ r_5 = [ 0.184 \ 0.359 \ -0.254]^{\tau} $ &  \\

 &  & $ r_6 = [ -0.22 \ -0.44 \ -0.04]^{\tau} $ &  \\

 &  & $ r_7 = [ 0.51 \ 0.79 \ -0.06]^{\tau} $ &  \\
\hline
$ {\multirow{7}{*}{ $ \hat{F}_i $}} $ & {\multirow{3}{*}{ }} & $ \hat{F}_1 = [ 0.08 \ 0.39 \ 0.92]^{\tau} $ & {\multirow{7}{*}{ $  $ }}  \\

  &  & $ \hat{F}_2 = [ -0.33 \ -0.90 \ 0.29]^{\tau} $ &  \\

  &  & $ \hat{F}_3 = [ 0.13 \ -0.87 \ -0.48]^{\tau} $ & \\

  & Thruster Orientations & $ \hat{F}_4 = [ 0.56 \ 0.08 \ 0.82]^{\tau} $ & \\

  &  & $ \hat{F}_5 = [ 0.83 \ 0.11 \ -0.55]^{\tau} $ & \\

  &  & $ \hat{F}_6 = [ -0.66 \ 0.57 \ -0.49]^{\tau} $ & \\

  &  & $ \hat{F}_7 = [ -0.59 \ 0.62 \ -0.51]^{\tau} $ & \\
\hline
\end{tabular}
\end{center}
\caption{Aerial Manipulator Parameters}
\label{table_parameters}
\end{table}
\section{Mathematical Model of the Aerial Manipulator} \label{model}
In this section, the kinematic and dynamic equations of motion in case there are no interaction forces and torques from the environment applied to the end-effector are presented.
\subsection{Kinematic Model}
Fig. \ref{fig:frames} shows the reference frames defined to derive the kinematic and dynamic model of the proposed system. The Earth-Fixed inertial frame is defined as $F_A = \{ \hat{x}_A, \hat{y}_A, \hat{z}_A \}$ and it should be noted that the Body-Fixed frame's origin does not coincide with the centre of gravity $ G $. The position of $ F_B $ relative to $ F_A $ can be represented by $ p = \left[ x \ y \ z \right]^{\tau} \in \mathbb{R}^3 $ and the corresponding orientation by the rotation angles $ \Theta = \left[ \phi \ \theta \ \psi \right]^{\tau} \in \mathbb{R}^3 $. The translational and rotational kinematic equations of the moving rigid body are given (see \cite{antonelli}) in matrix form by
\begin{equation} \label{eq:kin}
\dot{\xi} 
= 
\begin{bmatrix}
   \dot{p} \\
   \dot{\Theta}\\
\end{bmatrix}
=
\begin{bmatrix}
	J_t(\Theta) & O_{(3 \times 3)} \\
	O_{(3 \times 3)} & J_{ r }(\Theta) \\
\end{bmatrix}
\begin{bmatrix}
	v \\
	\omega \\
\end{bmatrix} \\
\end{equation}
where $O_{(3 \times 3)}$ is the $3 \times 3$ zero matrix, $ v = \left[ v_x \ v_y \ v_z \right] ^{\tau} \in \mathbb{R}^3
, \ \omega = \left[ \omega_x \ \omega_y \ \omega_z \right]^{\tau} \in \mathbb{R}^3 $ denote the translational velocity and the angular velocity of $ F_B $ relative to $ F_A $ respectively, both expressed in the Body-Fixed frame. The transformation matrices $ J_t(\Theta), J_r(\Theta) \in \mathbb{R}^{3 \times 3} $ are given by
\begin{equation} 
J_t(\Theta)=
\begin{bmatrix}
   c_{\theta} c_{\psi} & s_{\phi} s_{\theta} c_{\psi}-s_{\psi} c_{\phi} &
   s_{\theta} c_{\phi} c_{\psi}+s_{\phi} s_{\psi} \\
   s_{\psi} c_{\theta} & s_{\phi} s_{\theta} s_{\psi}+c_{\phi} c_{\psi} &
   s_{\theta} s_{\psi} c_{\phi}-s_{\phi} c_{\psi} \\
   -s_{\theta} & s_{\phi} c_{\theta} & c_{\phi} c_{\theta} \\
\end{bmatrix}
\end{equation}
\begin{equation}
J_r(\Theta) = 
\begin{bmatrix}
   1 & s_{\phi} t_{\theta}  & c_{\phi} t_{\theta} \\
   0 & c_{\phi} & -s_{\phi} \\
   0 & s_{\phi} / c_{\theta} & c_{\phi} / c_{\theta} \\
\end{bmatrix}
\end{equation}

The position of the end-effector with respect to $ F_A $ is $p_e = \left[ x_e \ y_e \ z_e \right]^{\tau} = p + J_t(\Theta) \ r_e \in \mathbb{R}^3$. Its derivative is obtained as $\dot{p}_e = J_t(\Theta) \ v - J_t(\Theta) \ S(r_e) \ \omega$, using the formula $\dot{J}_t(\Theta) = J_t(\Theta) S(\omega)$ from \cite{madani}. The Body-Fixed and the End-Effector frame have the same orientation with reference to $F_A$, as mentioned in Section \ref{sec2}, hence $\Theta_e = \Theta$. By combining the last results the following kinematic equation holds

\begin{equation} \label{eq:kin_ee}
\dot{\xi}_e 
= 
\begin{bmatrix}
   \dot{p}_e \\
   \dot{\Theta}\\
\end{bmatrix}
=
\underbrace{
\begin{bmatrix}
	J_t(\Theta) & -J_t(\Theta) \ S(r_e) \\
	O_{(3 \times 3)} & J_{ r }(\Theta) \\
\end{bmatrix} \\ }_{\substack{J(\xi_e)}}
\begin{bmatrix}
	v \\
	\omega \\
\end{bmatrix} \\
\end{equation}
where $\Theta_e$ is the orientation of $F_E$ relative to $F_A$. The Jacobian matrix of the system $ J(\xi_e) \in \mathbb{R}^{6 \times 6} $ relates in a straightforward way the linear velocity $\dot{p}_e$ and the rate of change in the rotational angles $\dot{\Theta}$ of the end-effector expressed in $ F_A $, with the Body-Fixed velocities $v,\omega$.
\subsection{Dynamic Model}
The dynamic equations can be conveniently written with respect to the Body-Fixed frame by using the Newton-Euler formalism (the main concept is discussed extensively in \cite{fossen, bernstein}), as
\begin{equation} \label{eq:dyn}
M \
\begin{bmatrix}
   \dot{v} \\
   \dot{\omega} \\
\end{bmatrix}
+C(\nu) \
\begin{bmatrix}
   v \\
   \omega \\
\end{bmatrix}
=
\begin{bmatrix}
   F \big \lvert_B \\
   T \big \lvert_B \\
\end{bmatrix}
\end{equation}
where
\begin{equation}
M = 
\begin{bmatrix}
   m I_{3} & -m S(r_G) \\
   m S(r_G) & I_B \\
\end{bmatrix}, \ M > 0, \ \dot{M}=0
\end{equation}
is the inertia matrix,
\begin{equation}
C = 
\begin{bmatrix}
   m S(\omega) & -m S(\omega) S(r_G) \\
   m S(r_G) S(\omega) & -S(I_B \omega) \\
\end{bmatrix}, C = -C^{\tau}
\end{equation}
is the Coriolis-centripetal matrix, $I_{3}$ is the $3 \times 3$ identity matrix, 
$ m $ is the total mass of the system, $ I_B $ is the inertia tensor expressed in $ F_B $ and $ \nu = \left[ v^{\tau} \ \omega^{\tau} \right]^{\tau} \in \mathbb{R}^{6} $ is the vector of the Body-Fixed velocities. The inertia tensor can be written as $ I_B = I_G-m S(r_G) S(r_G) $ where $ I_G $ is the inertia tensor relative to the body's centre of gravity. The vectors $ F \big \lvert_B, T \big \lvert_B \in \mathbb{R}^3 $ describe the forces and torques acting on the system expressed in the Body-Fixed frame and can be derived as
\begin{align}
\begin{bmatrix}
   F \big \lvert_B \\
   T \big \lvert_B \\
\end{bmatrix}
&=
\begin{bmatrix}
   F \ \lambda - m \ g \ J^{\tau}_t(\Theta) \ e_3  \\
   E \ \lambda + \mu \ F \ \lambda - m \ g \ S(r_G)\ J^{\tau}_t(\Theta) \ e_3  \\   
\end{bmatrix} \nonumber \\
&=
\underbrace{ 
\begin{bmatrix}
   F  \\
   E  \\   
\end{bmatrix} \lambda}_{\text{propulsion}\atop\text{forces/torques}}
+
\underbrace{ 
\begin{bmatrix}
   O_{( 3 \times 6)}  \\
   \mu \ F \\ 
\end{bmatrix} \lambda}_{\text{reaction}\atop\text{torques}}
-
\underbrace{
m \ g
\begin{bmatrix}
   I_{3} \\
   S(r_G) \\
\end{bmatrix} J_t^{\tau}(\Theta) e_3}_{\text{gravitational}\atop\text{forces/torques}} \label{eq:final2}
\end{align}
where $ e_3 = \left[ 0 \ 0 \ 1 \right]^{\tau} $. Combining  \eqref{eq:dyn}, \eqref{eq:final2} and solving with respect to $ \left[ \dot{v}^{\tau} \ \dot{\omega}^{\tau} \right]^{\tau} $ we get
\begin{align} 
\begin{bmatrix} 
   \dot{v} \\
   \dot{\omega} \\
\end{bmatrix}
&=
- M^{-1}  C(\nu) 
\begin{bmatrix}
   v \\
   \omega \\
\end{bmatrix}
 + M^{-1} \ 
\begin{bmatrix}
   F  \\
   E  + \mu \ F \\   
\end{bmatrix} \lambda
\nonumber \\
& \qquad {} - m \ g \ M^{-1} \
\begin{bmatrix}
   I_{3} \\
   S(r_G) \\
\end{bmatrix} J_t^{\tau}(\Theta) \ e_3 \label{eq:final_dyn1} \\
\Leftrightarrow
\dot{\nu} 
&=
\underbrace{ H(\nu) + G(\xi_e)}_{B(\xi_e,\nu)}
+N \ \lambda \label{eq:final_dyn2}
\end{align}
where the matrices are defined as
\begin{align}
& H(\nu) = -M^{-1} \ C(\nu) \ \nu, N = M^{-1} \ 
\begin{bmatrix}
   F  \\
   E + \mu \ F \\   
\end{bmatrix} > 0 \\
& G(\xi_e) = - m \ g \ M^{-1} \
\begin{bmatrix}
   I_{3} \\
   S(r_G) \\
\end{bmatrix} J_t^{\tau}(\Theta) e_3 \\
& B(\xi_e,\nu) = H(\nu) + G(\xi_e), \ B: \mathbb{R}^6 \times \mathbb{R}^6 \rightarrow \mathbb{R}^6 
\end{align}
\section{Nonlinear Control of the Aerial Manipulator} \label{control}
A manipulation task is usually given in terms of the desired position and orientation of the end-effector. The objective of this section is to design a controller for the aerial manipulator ensuring that the position $p_e(t)$ and the orientation $\Theta(t)$ of the end-effector track the desired Cartesian trajectory $ \xi_{\text{des}}(t) = \left[ p^{\tau}_{e,{\text{des}}}(t) \ \Theta^{\tau}_{\text{des}}(t) \right]^{\tau} \in \mathbb{R}^6 $ asymptotically while all the closed loop signals remain bounded for all $t \geq 0$. Firstly, by using formulas \eqref{eq:kin_ee}, \eqref{eq:final_dyn2} the aerial manipulator model, including the kinematics and dynamics, can be written as
\begin{equation} \label{eq:sys}
(S):
\begin{cases}
   \dot{\xi_e} = J(\xi_e) \ \nu \\
   \dot{\nu} = B(\xi_e,\nu)+N \ \theta^{\star}_{\lambda} \ \lambda+d(\xi_e,\nu,t) \\
\end{cases}
\end{equation}
where $d: \mathbb{R}^6 \times \mathbb{R}^6 \times \mathbb{R}_+ \rightarrow \mathbb{R}^6$ represents the unmodelled nonlinear dynamics and the environmental disturbances. The unknown matrix $\theta^{\star}_{\lambda} = \text{diag} \{ \theta^{\star}_{1},\dots, \theta^{\star}_{6} \} \in \mathbb{R}^{6 \times 6}$ with $\theta^{\star}_{i} \in [\theta_{\text{min}},\theta_{\text{max}}]=[0.1,1]$, is introduced to model the control actuation failures and the modeling errors among the thrusters of the system, e.g. if $\theta^{\star}_i = 0.8$ then the $i-$th actuator has 20 \% controller effectiveness reduction. The control inputs of the system are the six independent thrust forces $ \lambda_i(t), i = 1,\dots,6 $ as mentioned in Section \ref{sec2}. The matrix $N$ is full rank with low condition number which constitutes a vital result of the control oriented optimization from Section \ref{sec2}.  

The system \eqref{eq:sys} is highly nonlinear, cascaded and fully actuated in the well-known strict feedback form, with vector relative degree 2. For such systems, the backstepping controller design has proven to be successful \cite{kokotovic, zhou}. Due to the fact that the system is in the presence of the uncertainties $\theta^{\star}_{\lambda}$ and the disturbances $d(\xi_e,\nu,t)$, a robust adaptive controller will be designed in order to tackle them. The aim is to study if the proposed system with the resulting geometry from the optimization problems $(P_1),(P_2)$, the system specifications from Table \ref{table_parameters} and the aforementioned uncertainties/disturbances from \eqref{eq:sys}, is capable to perform specific trajectory tasks efficiently. In order to design the controller of the system \eqref{eq:sys}, the following assumptions are required:\\
\textit{\textbf{Assumption 1:}} The states of the system $ \xi_e, \nu $ are available for measurement $\forall t \geq 0$ for the following control development. \textit{\textbf{Assumption 2:}} The desired trajectories $ \xi_{\text{des}} $ are known and bounded functions of time ($ \xi_{\text{des}} \in \mathcal{L}_{\infty} $) with known and bounded derivatives ($ \dot{\xi}_{\text{des}}, \ddot{\xi}_{\text{des}} \in \mathcal{L}_{\infty} $). \textit{\textbf{Assumption 3:}} The disturbance $ d(\xi_e,\nu,t) = \left[ d_1(\xi_e,\nu,t) \ \cdots \ d_6(\xi_e,\nu,t) \right]^{\tau} $ is unknown but bounded with $|d_i(\xi_e,\nu,t)| \leq \Delta_i$ where $\Delta_i$ are unknown positive constants for all $i =1,\dots,6$ and $t \geq 0$. \textit{\textbf{Assumption 4:}} It is assumed for all $t \geq 0$ that $-\frac{\pi}{2} < \theta(t) < \frac{\pi}{2}$. This ensures that the Jacobian matrix is nonsingular since $ \text{det}(J(\xi_e)) = 1/c_\theta $. This assumption is likewise utilized in \cite{madani}, \cite{huang}. 

$ \bullet $ \textit{\textbf{Step 1:}} To begin with the backstepping controller design, the position-orientation error of the end-effector is defined as $z_1 = \xi_e-\xi_{\text{des}} \in \mathbb{R}^6$. By differentiating it and using \eqref{eq:kin_ee} we get
\begin{equation} \label{eq:error2}
\dot{z}_1 = J(\xi_e) \ \nu - \dot{\xi}_{\text{des}}
\end{equation}
We view $ \nu $ as a control variable and we define a virtual control law $ \nu_{\text{des}} \in \mathbb{R}^6 $ for \eqref{eq:error2}. The error signal representing the difference between the virtual and the actual controls is defined as $z_2 = \nu-\nu_{\text{des}} \in \mathbb{R}^6$. Thus, in terms of the new state variable, \eqref{eq:error2} can be rewritten as $\dot{z}_1 = J(\xi_e) \ z_2 + J(\xi_e)  \ \nu_{\text{des}}-\dot{\xi}_{\text{des}}$. Consider now the positive definite and radially unbounded quadratic Lyapunov function $V_1(z_1) = \frac{1}{2} \| z_1 \|^2 = \frac{1}{2}z_1^{\tau}z_1$. By differentiating it with respect to time yields
\begin{equation}
\dot{V}_1 = z_1^{\tau} \dot{z}_1 = z_1^{\tau} \left\lbrace J(\xi_e) \ \nu_{\text{des}} -\dot{\xi}_{\text{des}} \right\rbrace+z_1^{\tau} J(\xi_e) z_2
\end{equation}
The stabilization of $ z_1 $ can be obtained by designing an appropriate virtual control law
\begin{equation} \label{eq:vdes}
\nu_{\text{des}} = J^{-1}(\xi_e) \ \left\lbrace \dot{\xi}_{\text{des}}-K_1 z_1 \right\rbrace 
\end{equation}
where the matrix $K_1 \in \mathbb{R}^{6 \times 6}, \displaystyle K_1 = K^{\tau}_1 > 0$ represents the first controller gain to be designed. Hence, the time derivative of $ V_1 $ becomes $\dot{V}_1 = -z_1^{\tau} K_1 z_1 +z_1^{\tau} J(\xi_e) z_2$. The first term of on the right-hand of this equation is negative and the second term will be canceled in the next step. 

$\bullet$ \textit{\textbf{Step 2:}} For the second step we define the matrices of the parameter estimation errors as $\tilde{\Delta} = [ \tilde{\Delta}_1 \cdots \tilde{\Delta}_6 ]^{\tau} = [ (\hat{\Delta}_1-\Delta_1) \cdots (\hat{\Delta}_6-\Delta_6) ]^{\tau}$ and $\tilde{\theta}_{\lambda} = \text{diag} \{(\hat{\theta}_1-\theta^{\star}_1),\dots, (\hat{\theta}_6-\theta^{\star}_6) \}$ where $\hat{\Delta}_i, \hat{\theta}_i$ are the estimations of the unknown parameters $\Delta_i, \theta^{\star}_i$ respectively. The time derivative of the error $z_2$ is $\dot{z}_2 = B(\xi_e,\nu)+N \ \theta^{\star}_{\lambda} \ \lambda+d(\xi_e,\nu,t)-\dot{\nu}_{\text{des}}$. The Lyapunov function candidate in this step is chosen as
\begin{equation}
V_2(z_1,z_2,\tilde{\Delta},\tilde{\theta}_{\lambda}) = V_1+\frac{1}{2} z_2^{\tau}z_2+\frac{1}{2} \tilde{\Delta}^{\tau} \Gamma_{\Delta}^{-1} \tilde{\Delta}+\frac{1}{2} \text{tr} (\tilde{\theta}_{\lambda}^{\tau} \Gamma_{\theta}^{-1} \tilde{\theta}_{\lambda} ) \nonumber
\end{equation}
where $ \displaystyle \Gamma_{\theta} = \Gamma_{\theta}^{\tau} > 0, \displaystyle \Gamma_{\Delta} = \Gamma_{\Delta}^{\tau} > 0$ are diagonal adaptation gain matrices and $\text{tr}(\cdot)$ denotes the matrix trace. The time derivative of $V_2(z_1,z_2,\tilde{\Delta},\tilde{\theta}_{\lambda})$ is obtained as
\begin{align}
\dot{V}_2 &= -z_1^{\tau} K_1 z_1 + z_2^{\tau} \left\lbrace  J^{\tau}(\xi_e) z_1 + B(\xi_e,\nu)+N \ \theta^{\star}_{\lambda} \ {\lambda} -\dot{\nu}_{\text{des}} \right\rbrace \nonumber \\
& \qquad +z_2^{\tau} d(\xi_e,\nu,t) +\tilde{\Delta}^{\tau} \ \Gamma_{\Delta}^{-1} \dot{\hat{\Delta}} + \text{tr} (\tilde{\theta}_{\lambda}^{\tau} \Gamma_{\theta}^{-1} \dot{\hat{\theta}}_{\lambda} ) \label{eq:lyap22}
\end{align}
Using the $-z_1^{\tau} K_1 z_1 \leq -\lambda_{\text{min}} (K_1) \| z_1\|^2, \ z_2^{\tau} \ d(\xi_e,\nu,t) \leq z_2^{\tau} \text{sgn}(z_2) \Delta $ and adding and subtracting the terms $z_2^{\tau} N \hat{\theta}_{\lambda} {\lambda}, \ z_2^{\tau} \text{sgn}(z_2) \hat{\Delta}$ in \eqref{eq:lyap22} the following inequality holds
\begin{multline}
\dot{V}_2 \leq -\lambda_{\text{min}}(K_1)\|z_1\|^2 + z_2^{\tau} \big\{ J^{\tau}(\xi_e) z_1 + B(\xi_e,\nu)-\dot{\nu}_{\text{des}} \\
+ \text{sgn}(z_2) \ \hat{\Delta} +N \ \hat{\theta}_{\lambda} {\lambda} \big\} -z_2^{\tau} N \tilde{\theta}_{\lambda} {\lambda}  \\ 
-z_2^{\tau} \text{sgn}(z_2) \ \tilde{\Delta}+\tilde{\Delta}^{\tau} \ \Gamma_{\Delta}^{-1} \dot{\hat{\Delta}}+\text{tr} (\tilde{\theta}^{\tau}_{\lambda} \Gamma_{\theta}^{-1} \dot{\hat{\theta}}_{\lambda} )
\end{multline}
where $\lambda_{\text{min}} (K_1)$ denotes the minimum eigenvalue of matrix $K_1$, $\text{sgn}(z_2) = \text{diag} \{ \text{sgn}(z_{2,1}), ..., \text{sgn}(z_{2,6}) \}$ and $\text{sgn}(\cdot)$ denotes the sign function. Rearranging the terms and using the property $a^{\tau}b = \text{tr}(b \ a^{\tau}), \ \forall a,b \in \mathbb{R}^n$ we get
\begin{multline}
\dot{V}_2 \leq -\lambda_{\text{min}}(K_1)\|z_1\|^2 + z_2^{\tau} \big\{ J^{\tau}(\xi_e) z_1 + B(\xi_e,\nu)-\dot{\nu}_{\text{des}} \\
+ \text{sgn}(z_2) \ \hat{\Delta} +N \ \hat{\theta}_{\lambda} \ {\lambda} \big\}   
+\tilde{\Delta}^{\tau} \big\{\Gamma_{\Delta}^{-1} \dot{\hat{\Delta}}
- \text{sgn}(z_2)  z_2 \big\} \\ 
+\text{tr} \{ \tilde{\theta}_{\lambda}^{\tau} ( \Gamma_{\theta}^{-1}\dot{\hat{\theta}}_{\lambda}- N^{\tau} z_2 \ {\lambda}^{\tau}  ) \} \label{eq:lyap_fin}
\end{multline}

Given the form of $\dot{V}_2$ from \eqref{eq:lyap_fin} the adaptive control law and the corresponding parameter estimation update laws for the nonlinear system \eqref{eq:sys} to be designed, are
\begin{align} 
\lambda(\xi_e,\nu,\hat{\Delta},\hat{\theta}_{\lambda}) &= (\hat{\theta}_{\lambda})^{-1} N^{-1} \big\{  \dot{\nu}_{\text{des}}-B(\xi_e,\nu)-J^{\tau}(\xi_e) z_1 \nonumber \\
& \qquad \ \ -\text{sgn}(z_2) \ \hat{\Delta}-K_2 z_2 \} \label{eq:control} \\
\dot{\hat{\Delta}} &= \Gamma_{\Delta} \{ \text{sgn}(z_2)  z_2 -\sigma \ \hat{\Delta} \} \label{eq:adapt1} \\
\dot{\hat{\theta}}_{\lambda} &= \Gamma_{\theta} \ \text{Proj}(\hat{\theta}_{\lambda},N^{\tau} z_2 \ {\lambda}^{\tau} ) \label{eq:adapt2}
\end{align}
where $\displaystyle K_2 = K_2^{\tau} > 0$ is the second controller gain matrix, $\sigma$ is a strictly positive gain ($\sigma$-modification rule \cite{lavretsky}) and the projection operator $\text{Proj}(\cdot,\cdot)$ is the same as the one in \cite{khalil} with the parameter $\delta$ to be designed. By substituting \eqref{eq:control}, \eqref{eq:adapt1}, \eqref{eq:adapt2} into \eqref{eq:lyap_fin} and using the property $ \displaystyle \tilde{\Delta}^{\tau} \hat{\Delta} = \frac{1}{2} \| \tilde{\Delta}\|^2+\frac{1}{2} \|\hat{\Delta}\|^2-\frac{1}{2} \| \Delta \|^2$ the following inequality holds
\begin{multline}
\dot{V}_2 \leq -\lambda_{\text{min}}(K_1)\|z_1\|^2 -\lambda_{\text{min}}(K_2)\|z_2\|^2 + \\
\underbrace{ \text{tr} \left\{ \tilde{\theta}_{\lambda}^{\tau} \left[ \text{Proj}(\hat{\theta}_{\lambda},y )-y \right] \right\} }_{ \displaystyle \leq 0, \  y = N^{\tau} z_2 \ {\lambda}^{\tau} } \underbrace{ -\frac{\sigma}{2} \| \tilde{\Delta}\|^2-\frac{\sigma}{2} \|\hat{\Delta}\|^2+\frac{\sigma}{2} \| \Delta \|^2}_{ \leq -\frac{\sigma}{2} \| \tilde{\Delta}\|^2+\frac{\sigma}{2} \| \Delta \|^2} \nonumber
\end{multline}
The projection operator invoked from \cite{khalil} contributes to the negative semi-negativeness of the Lyapunov function since by definition $\text{tr} \left\{ \tilde{\theta}_{\lambda}^{\tau} \left[ \text{Proj}(\hat{\theta}_{\lambda},y )-y \right] \right\} \leq 0, \ \forall y$. Moreover, it guarantees that if $\hat{\theta}_i(0) \in [ \theta_{\text{min}}, \theta_{\text{max}} ]$ is chosen, then $\hat{\theta}_i(t) \in [ \theta_{\text{min}}-\delta, \theta_{\text{max}}+\delta ], \ \forall i=1,...,6, \ \forall t \geq 0$ for suitable $\delta>0$. The last result protects the term $(\hat{\theta}_{\lambda})^{-1}$ in \eqref{eq:control} from singularity. By defining $\bar{w} = \frac{\sigma}{2} \| \Delta \|^2 > 0$ we result in
\begin{multline}
\dot{V}_2 \leq -\lambda_{\text{min}}(K_1)\|z_1\|^2 -\lambda_{\text{min}}(K_2)\|z_2\|^2 -\frac{\sigma}{2} \| \tilde{\Delta}\|^2 + \bar{w} \nonumber
\end{multline} 
from which it follows that both errors $z_1,z_2$ and the parameter estimation $\tilde{\Delta}$ are uniformly ultimately bounded with respect to the sets $\Omega_1 = \left\lbrace  z_1 \in \mathbb{R}^6 : \| z_1 \| \leq \sqrt{\bar{w} / \lambda_{\text{min}}(K_1)} \right\rbrace$, $\Omega_2 = \left\lbrace  z_2 \in \mathbb{R}^6 : \| z_2 \| \leq \sqrt{\bar{w} / \lambda_{\text{min}}(K_2)} \right\rbrace$ and $\Omega_{\Delta} = \left\lbrace  \tilde{\Delta} \in \mathbb{R}^6 : \| \tilde{\Delta} \| \leq \sqrt{2 \ \bar{w} / \sigma} \right\rbrace$. Invoking that $z_1,z_2$ are bounded and $\xi_{\text{des}}, \nu_{\text{des}} \in \mathcal{L}_{\infty} $ then $ \xi_e, \nu \in \mathcal{L}_{\infty} $. Since $\tilde{\Delta}, \Delta, \hat{\theta}_{\lambda}, \theta^{\star}_{\lambda}, \dot{\nu}_{\text{des}}$ are bounded then $ \hat{\Delta}, \ \tilde{\theta}_{\lambda}, \lambda \in \mathcal{L}_{\infty} $. Based on the above, it is proven that all closed loop signals remain bounded.

One important issue associated with the controller design is the analytical form of the time derivative of $\nu_{\text{des}}$, which can be obtained from \eqref{eq:vdes} as $\dot{\nu}_{\text{des}} = J^{-1}(\xi_e) \left\lbrace \ddot{\xi}_{\text{des}} - \dot{J}(\xi_e) \ \nu_{\text{des}} - K_1 \ \dot{z}_1 \right\rbrace $, and the time derivative of $ J(\xi_e) $, which can be calculated by using the $\dot{J}_r(\Theta) = \displaystyle \frac{\partial J_r}{\partial \phi} \ \dot{\phi} + \displaystyle \frac{\partial J_r}{\partial \theta} \ \dot{\theta}$, $ 
\dot{J}(\xi_e) = 
\begin{bmatrix}
	\dot{J}_t(\Theta) & -\dot{J}_t(\Theta) S(r_e) \\
	O_{(3 \times 3)} & \dot{J}_{ r }(\Theta) \\
\end{bmatrix}$.
\begin{figure}[h]
\centering
\includegraphics[width=65mm,height=41mm]{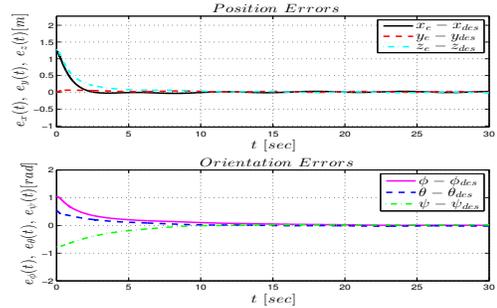}
\caption{Position and Orientation Errors}
\label{fig:pos_errors}
\end{figure} 
\begin{figure}[h]
\centering
\includegraphics[width=65mm,height=41mm]{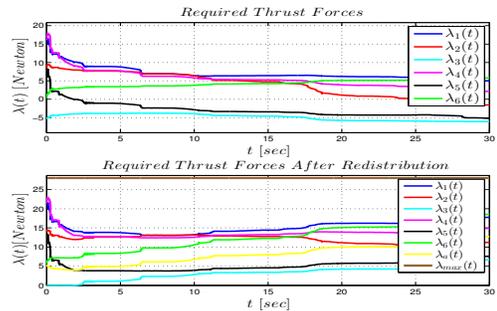}
\caption{Required Thrust Forces Using the Algorithm \eqref{eq:final proof}}
\label{fig:redistribution}
\end{figure}
\section{Simulation Results} \label{simul}
In this section, the results of a numerical simulation scenario are presented in order to demonstrate the performance of the proposed system. The dynamic model in \eqref{eq:sys} is utilized with system parameters which are depicted in Table \ref{table_parameters}. 15 \% controller effectiveness reduction is chosen with $\theta^{\star} = 0.85 \ \text{diag} \{1,1,1,1,1,1\}$. The end-effector is forced to track the trajectory $p_{\text{e,des}}(t) = \left[ \cos(0.5t) \ \sin(0.5t) \ 1.5+0.3t \right]^{\tau}$ with regulated orientation at $\Theta_{\text{des}} = \left[ \frac{\pi}{3} \ \frac{\pi}{6} \ -\frac{\pi}{4} \right]^{\tau}$ with reference to the Earth-Fixed frame. The initial conditions of the system are set to $p_{\text{e}}(0) = r_e, \ p(0) = \Theta(0) = v(0) = \omega(0) = 0_{(3 \times 1)}$. The disturbance is set as $d(t) = [0.5 \ 0.4 \sin(2t) \ 0.4 \cos(t) \ 0.5 \ 0.5 \cos(0.8t) \ 0.6 \sin(t) ]^{\tau}$ representing the unmodelled forces/torques and external disturbances. The initial values of the parameters estimations are set to $\hat{\theta}_1(0) =\dots= \hat{\theta}_6(0) = 0.7$ and $\hat{\Delta}_1(0) = \dots = \hat{\Delta}_6(0) = 0$. The controller gains are chosen as $ K_1 = \text{diag} \{ 1, 1, 1, 0.3, 0.3, 0.3 \}, K_2 = 8 \ \text{diag} \{ 1, 1, 1, 1, 1, 1 \} $. The adaptation gains are selected as $\sigma = 1.5, \ \Gamma_{\Delta} = 13 \ \text{diag} \{ 1,1,1,1,1,1 \}, \Gamma_{\theta} = 0.1 \ \text{diag} \{ 1,1,1,1,1,1 \} $. The parameter of the projection operator is set to $\delta = 0.05$. Fig. \ref{fig:pos_errors} shows the position and orientation tracking errors. The thrust forces are provided in Fig. \ref{fig:redistribution}. This paper is accompanied by a video demonstrating the simulation procedure of this Section. Due to space limitations, a video with an additional Scenario in better quality (HD) can be found at

{ \tt \small https://www.youtube.com/watch?v=DXnzu6XOrXs }
\section{Conclusions} \label{conclusions}
Aerial robots physically interacting with the environment could be very useful for many applications. In this paper, we have presented the mechanical design of a novel aerial manipulator which was the result of technical optimization problems. A mathematical model for the kinematics and dynamics was derived in order to design an adaptive nonlinear controller to study the system while performing manipulation tasks. The simulation results illustrate the effectiveness of the proposed system and the controller to achieve tracking irrespectively of actuator failures, unmodelled dynamics and external disturbances. Future work mainly involves the construction of the aerial robot and the conduction of experimental trials for the proposed framework with the actual system, in order to verify the theoretical results of this paper.
\bibliography{bibfile}
\bibliographystyle{ieeetr}
\addtolength{\textheight}{-12cm}  
\end{document}